\documentclass{svjour3}

\usepackage{array}
\usepackage[table,xcdraw]{xcolor}
\usepackage{url}
\usepackage{amsmath,graphicx}
\usepackage{times}
\usepackage{booktabs}
\usepackage{tabularx}
\usepackage{multirow}
\usepackage{fancyhdr}
\usepackage{rotating}
\usepackage{bm}
\usepackage{epstopdf}
\usepackage{hyperref}
\usepackage{tabularx}
\usepackage{multirow}
\usepackage{rotating}

\newcolumntype{x}[1]{>{\centering\hspace{0pt}}p{#1}}

\smartqed
\usepackage{amssymb,amsmath}

\journalname{Artificial Intelligence Review}
\begin{document}

\title{Recent Trends in Deep Learning Based Personality Detection}
\author{Yash Mehta \and Navonil Majumder \and Alexander Gelbukh \and Erik Cambria}
\institute{Y.~Mehta\at
Gatsby Computational Neuroscience Unit,\\
University College London\\\\
N.~Majumder and A.~Gelbukh\at
Centro de Investigaci\'on en Computaci\'on,\\
Instituto Polit\'ecnico Nacional\\\\
E.~Cambria\at
School of Computer Science and Engineering,\\
Nanyang Technological University, Singapore\\
\email{\texttt{cambria@ntu.edu.sg}}}

\date{Received: 2018-12-03 / Accepted: XXXX-XX-XX}
\maketitle
\begin{abstract}
Recently, the automatic prediction of personality traits has received a lot of attention. Specifically, personality trait prediction from multimodal data has emerged as a hot topic within the field of affective computing. In this paper, we review significant machine learning models which have been employed for personality detection, with an emphasis on deep learning-based methods. This review paper provides an overview of the most popular approaches to automated personality detection, various computational datasets, its industrial applications, and state-of-the-art machine learning models for personality detection with specific focus on multimodal approaches. Personality detection is a very broad and diverse topic: this survey only focuses on computational approaches and leaves out psychological studies on personality detection.
\end{abstract}


%
%

\section{Introduction}

Personality is the combination of an individual's behavior, emotion, motivation, and characteristics of their thought
patterns. Our personality has a great impact on our lives, affecting our life choices, well-being, health along with
our preferences and desires. Hence, the ability to automatically detect one's personality traits has many important practical applications. The Woodworth Psychoneurotic Inventory~\cite{papurt1930study} is commonly cited as the first personality test. It was developed during World War I for the U.S. military to screen recruits for shell shock risks (post-traumatic stress disorders).
In the present day, a personality model which has an active community and is used by many is the Process Communication Model (PCM). It was developed by Taibi Kahler with NASA funding and was initially used to assist with shuttle astronaut selection. This measure is now being used mainly for consultancy purposes in order to help individuals to become more effective in communication and even avoid or resolve situations where communication has gone off track.
\par
Instead of directly determining an individual's personality, one may also be interested in
knowing how they are perceived by the people around them. Unlike the case of automatic personality recognition, the target of perceived personality analysis is not the true personality
of individuals, but the personality which is attributed to them
by the people who interact with them. The surrounding people fill a similar personality questionnaire on the individual, which then determines the perceived personality of that person. The perceived personality is evaluated using the same personality measures used for their `true' personality (e.g., Big-Five). A point to be noted is that the perceived personality varies from person to person and is
subject to their mental make or their own personality.
Many studies have shown that people form a first impression about a person by their facial
characteristics within the first ten seconds of meeting an individual. Apart from the physical characteristics of
the encountered face, the emotional expressions, the ambient information, the psychological formation of the viewer also affect these impressions. First impressions influence the behavior of people towards a newly encountered person or a human-like agent.

\subsection{Personality Measures}
Modern trait theory~\cite{pervin1999handbook} tries to model personality by setting of a number of classification dimensions (usually following a lexical
approach) and constructing a questionnaire to measure them~\cite{matthews2003personality}.
Researchers have used various schemes for personality modeling such as 16PF~\cite{cattell2008sixteen}, EPQ-R
\cite{miles2004eysenck} and three trait personality model PEN~\cite{eysenck2012model} where there are superfactors Psychoticism, Extraversion, and Neuroticism (PEN) at the top of the hierarchy. The Myers--€"Briggs Type Indicator (MBTI)~\cite{briggs1993introduction} is one of the most widely administered personality test in the world, given
millions of times a year to employees in thousands of companies. The MBTI personality measure categorizes people
into two categories in each four dimensions: introversion versus extraversion, sensing versus intuiting, thinking versus
feeling, and judging versus perceiving.

The most popular measure used in the literature on automated personality detection is by far the Big-Five personality traits~\cite{digman1990personality}, which are the following binary (yes/no) values:

\begin{itemize}

  \item{\textbf{Extraversion (EXT)}: Is the person outgoing, talkative, and energetic versus reserved and solitary?}
  \item{\textbf{Neuroticism (NEU)}: Is the person sensitive and nervous versus secure and confident?}
  \item{\textbf{Agreeableness (AGR)}: Is the person trustworthy, straightforward, generous, and modest versus unreliable, complicated, meager and boastful?}
  \item{\textbf{Conscientiousness (CON)}: Is the person efficient and organized versus sloppy and careless?}
  \item{\textbf{Openness (OPN)}: Is the person inventive and curious versus dogmatic and cautious?}

\end{itemize}
\par
As majority of the research makes use of the Big-Five personality measure for classification, we will refer to this measure by default, unless stated otherwise.

\subsection{Applications} \label{Applications}
There are various diverse industrial applications of automated personality recognition systems in the present day scenario. We surmise that a huge market will open up in the near future and if models are able to measure personality accurately and consistently, there will be a huge demand for automated personality recognition software in the industry. As research progresses in this field, soon better personality prediction models with much higher accuracies and reliability will be discovered. Artificial personality can be integrated with almost all human computer interactions going forward. 

Any computational devices can be equipped with some sort of personality which enables it to react differently to different people and situations. For example, a phone can have different modes with different configurable personalities. This will pave way for more interesting and personalized interactions.
Another possibility is to use personality traits as one of the inputs for achieving higher accuracy in other tasks such as sarcasm detection, lie detection or word polarity disambiguation systems.
\begin{itemize}
    \item{\textit{\textbf{Enhanced Personal Assistants}}}: Present day automated voice assistants such as Siri, Google Assistant, Alexa, etc. can be made to automatically detect personality of the user and, hence, give customized responses. Also, the voice assistants can be programmed in such a way that they display different personalities based on the user personality for higher user satisfaction.

  \item{\textit{\textbf{Recommendation systems}}: People that share a particular personality type may have similar interests and hobbies. The products and services that are recommended to a person should be those that have been positively evaluated by other users with a similar personality type. For example, \cite{yin2018network} propose to model the automobile purchase intentions of customers based on their hobbies and personality. \cite{yang2019mining} have developed a system for recommending games to players based on personality which is modelled from their chats with other players.}

 \item{\textit{\textbf{Word polarity detection}}: Personality detection can be exploited for word polarity disambiguation in sentiment lexicons, as the same concept can convey different meaning to different types of people. Also, incorporating user personality traits and profile for disambiguation between sarcastic and non-sarcastic content~\cite{poria2016deeper} gives improved accuracy results.}

 \item{\textit{\textbf{Specialized health care and counseling}}: As of 2016, nearly one-third of Americans have sought professional counseling for mental health related issues. This is yet another area where huge practical applications of personality trait prediction exists. According to an individual's personality, appropriate automated counseling may be given or a psychiatrist may make use this information to give better counseling advice.}

 \item{\textit{\textbf{Forensics}}:  If the police are aware of the personality traits of the people who were present at the crime scene, it may help in reducing the circle of suspects. Personality detection also helps in the field of automated deception detection and can help in building lie detectors with higher accuracies.}

 \item{\textit{\textbf{Job screening}}: In human resource management, personality traits affect one's suitability for certain jobs. For example, maybe a company wants to recruit someone who will motivate and lead a particular team. They can narrow down their screening by eliminating candidates who are highly nervous and sensitive, i.e., those having high values of neuroticism trait. \cite{liem2018psychology} discusses the job candidate screening problem from an interdisciplinary viewpoint of psychologists and machine learning scientists.}

 \item{\textit{\textbf{Psychological studies}}: Automated personality trait detection will help find more complex and subtle relations between people's behaviorism and personality traits. This will aid in discovering new dynamics of the human psyche.}

 \item{\textit{\textbf{Political forecasting}}}:
Large scale automated personality detection is being used as a guideline for politicians to come up with more effective and targeted campaigns. If an analytical firm is able to procure large scale behavioral data about the voters, the firm can then create their psychographical profiles. These profiles can give an insight to the kind of advertisement that would be most effective in persuading a particular person at a particular location for some political event.
\end{itemize}

\subsection{Fairness and Ethics}
Over the past century, various techniques to access personality have been developed and refined by the practitioners and researchers in the field. Several ethical guidelines have been set to ensure that proper procedures are followed. However, often, personality detection involves ethical dilemmas regarding appropriate utilization and interpretations~\cite{mukherjee2016ethical}. Concerns have been raised regarding the inappropriate use of these tests with respect to invasion of privacy, cultural bias and confidentiality.

Recently, political parties are trying to leverage large-scale machine learning based personality detection for political forecasting. A rather infamous example showing the use of personality detection in political forecasting is the Facebook's Cambridge Analytica data scandal\footnote{The Great Hack, a recent documentary about the Cambridge Analytica data scandal}, which involved the collection of personally identifiable information of 87 million Facebook users. Studies have shown that the political choices of voters have a strong correlation with their social characteristics~\cite{caprara2006personality}. Users' social data were allegedly used in an attempt to influence voter opinion in favor of the 2016 Trump election campaign by identifying persuasive advertisement types effective for a particular region. It also involved automatically accessing personality traits of voters and using this to predict  who would be manipulated easily by political propaganda.

There is also concern regarding the continued use of these tests despite them lacking valid proof of being accurate measures of personality. These tests may have inherent biases, which are then inadvertently learned by the machine learning algorithm. One step to reduce biases in machine learning algorithms is through Algorithmic Impact Assessments (AIAs), as proposed by New York University's AI Now Institute. AIAs extend from the idea that the "black box" methodology leads to a vicious cycle, continuously moving further away from understanding these algorithms and diminishing the ability to address any issues that may arise from them. AI Now suggests the use of AIAs to handle the use of machine learning in the public realm, creating a set of standard requirements. Through AIAs, AI Now aims to provide clarity to the public by publicly listing and explaining algorithm systems used while allowing the public to dispute these systems, developing an audit and assessment processes, and increasing public agencies' internal capabilities to understand the systems they use. Similar to the use of AIAs to promote transparency in machine learning, the Defense Advanced Research Projects Agency (DARPA) suggests Explainable Artificial Intelligence (XAI) as a part of the solution. The goal is to produce more explainable models that users can understand and trust.

\subsection{Organization}
\par
We have thoroughly reviewed machine learning-based personality detection, with focus on deep learning-based approaches. 
Section~\ref{Related Works} gives the reader an idea of other review papers on machine learning-based automated personality detection.
We have divided our analysis based on the modality of the input, such as text, audio, video, or multimodal. Section~\ref{Baseline Methods} briefly discusses the popular techniques and methods used for personality detection in each of these modalities. In Section~\ref{Detailed Overview}, we analyze multiple interesting papers in-depth in each of these modalities. Results with analysis and conclusions follow in Section~\ref{Results and Discussions} and Section~\ref{Conclusion}, respectively.

\begin{table*}[t]
\centering
\caption{The various popular tools used for feature extraction for each of the modalities.}

\begin{tabular}{@{}cll@{}}
\toprule

\rowcolor[HTML]{EFEFEF}
\multicolumn{1}{l}{\cellcolor[HTML]{EFEFEF}Modality} & Tool & Features Extracted \\
\midrule
\rowcolor[HTML]{ECF4FF}
Text                                                 & \begin{tabular}[c]{@{}l@{}}LIWC\\ Receptiviti API\\ Freeling\\ SenticNet\end{tabular} & \begin{tabular}[c]{@{}l@{}}Psychological and linguistic features\\ Built on top of LIWC. Visual representation of personality traits\\ POS tagging\\ Sentiment polarity\end{tabular}                                                                                                \\
\rowcolor[HTML]{FFFFC7}
Audio                                                & \begin{tabular}[c]{@{}l@{}}Praat\\ OpenSMILE\end{tabular}                             & Intensity, pitch, loudness, MFCC, jitter, shimmer, LSP                                                                                                                                                                                                                              \\
\rowcolor[HTML]{FFCCC9}
Visual                                               & \begin{tabular}[c]{@{}l@{}}FACS\\ CERT\\ Face ++\\ EmoVu\end{tabular}                 & \begin{tabular}[c]{@{}l@{}}Facial Expressions, AUs\\ 8 facial expressions (e.g., joy, sadness, surprise, etc.)\\ Face detection, 106 facial landmarks, face attributes (e.g., gender, age)\\ Face detection, emotion recognition (with intensity), face attributes\end{tabular} \\ \bottomrule
\end{tabular}
\label{table1}
\end{table*}

\section{Related Works} \label{Related Works}
Automated personality detection is a new and upcoming field. There have not been many comprehensive literature surveys done in personality detection and our paper is the first one which gives the reader a bird's-eye view of the recent trends and developments in the field.

There is no recent work which gives the reader an overall perspective of the advances in machine learning based automated personality detection. 
Section~\ref{Baseline Methods} gives the popular machine learning models for each modality, which gives the reader a wide perspective in this field.
A review of the trends in personality detection from text using shallow learning techniques such as Na\"ive Bayes, kNN, mLR, Gaussian Process is given by~\cite{agarwal2014personality}. Post 2014, end-to-end deep neural network architectures and models picked up and started beating the state-of-the-art accuracies of these models. A review of the the various image processing techniques and facial feature analysis for personality detection is given by~\cite{Ilmini2016PersonsP}, in which their focus lies on an overview of machine learning techniques along with various facial image preprocessing techniques. 

Popular image preprocessing techniques in this domain include facial landmark identification, Facial Action Coding System (FACS), which gives the AUs in the face, and the Viola Jones face detection algorithm~\cite{viola2004robust}. Also, instead of using a standard measure of personality (such as Big-Five or MBTI), they have used slightly different metrics such as unhappy, weird, intelligent, confident, etc. This survey only focuses on the visual aspect whereas~\cite{vinciarelli2014survey} gives a very systematic and clear survey of personality detection from text as well as audio. Instead of dividing personality detection by modality as it is done in this paper, they have chosen a broader topic which they call personality computing. This is further divided into 3 fundamental problems, namely Automatic Personality Perception (APP), Automatic Personality Synthesis (APS) and Automatic Personality Recognition (APR). This paper focuses on methods following APR and APP (similar to perceived personality). The paper cited above gives a very detailed insight of the different papers dealing with recognition and perception of personality from social media, mainly Twitter and Facebook.
They present a thorough analysis of personality recognition from text, non-verbal communication (e.g., interpersonal distances, speech and body movements), social media, mobiles, wearable devices and finally from computer games as well.
A recent paper, \cite{junior2018first} provides a comprehensive survey on computer vision-based perceived personality trait analysis.
\cite{escalera2018guest} have made a compilation of the latest progress on automatic analysis of apparent personality in videos and images from multimedia information processing, pattern recognition and computer vision points of view.
However, there are still several deep learning techniques, models and papers, especially for multimodal personality detection which have not been covered by any literature survey till date.

\section{Baseline Methods} \label{Baseline Methods}
The following subsections summarize the various popular models, architectures and techniques commonly used in deep learning-based personality detection. Table~\ref{table3} provides a brief insight into the methods described by some of the pioneering papers in the field.

\subsection{Text}

Pertaining to the textual modality, data preprocessing is a very important step and choosing the correct technique can yield significantly better results. Usually, features from text are extracted such as Linguistic Inquiry and Word Count (LIWC)~\cite{pennebaker2001linguistic}, Mairesse, Medical Research Council (MRC), etc. which are then fed into standard ML classifiers such as Sequential Minimum Optimizer, Support Vector Machine (SVM), Na\"ive Bayes, etc. Learning word embeddings and representing them as vectors (with GloVe or Word2Vec) is also a very commonly followed approach. These word vectors may also be created by feeding the word character-wise into a Long Short-Term Memory (LSTM) or a Gated Recurrent Unit (GRU).
It was observed that combining text features (LIWC, MRC) with something else such as commonsense knowledge, convolutions, etc. results in better performance.
\subsection{Audio}

There are relatively few methods which focus on using only audio as the sole input for detecting personality. It is usually combined with visual modality for bimodal personality detection. Standard audio features such Mel-Frequency Cepstral Coefficients (MFCC), Zero Crossing Rate (ZCR), Logfbank, other cepstral and spectral ones serve as inputs into SVM and linear regressors.
\subsection{Visual}

As in the case of most visual deep learning tasks, Convolutional Neural Networks (CNN) are the most commonly used and yield state-of-the-art results in the field of personality detection as well. Most approaches analyze facial features and try to find a function which maps these features to personality traits. Many researchers have used pretrained deep CNNs (such as VGG-Face) and fine tuned it for the task of personality detection (transfer learning). They have experimented with different ways of extracting facial features such as EigenFace, Histogram of Oriented Gradients (HOG)~\cite{dalal2005histograms}, FACS (extracts AUs such as raised eyebrows, dimples, etc.) and Viola Jones Algorithm (face identification) to achieve higher accuracy.

\subsection{Multimodal}

Most of the multimodal approaches perform late fusion, that is, they take the average of individual results of predictions from the audio and visual modalities. Deep bimodal regression give state-of-the-art results by making use of slightly modified Deep Residual Networks~\cite{he2016deep}.
The features extracted from each of the modalities may be used together to come up with the personality prediction. This technique is early fusion.
Present research in the field takes the direction of finding more efficient ways of feature extraction along with multimodal feature combination. We see very few models which have dealt with trimodal fusion of features.

\section{Detailed Overview} \label{Detailed Overview}

In this section, we analyze individual papers in detail, discussing various techniques, approaches and methods. On an average, we find that features extracted from the visual modality are most accurate in unimodal personality detection. Studies have found that combining inputs from more than one modality often results in a higher prediction accuracy, as one may expect.
\subsection{\textbf{Text}}
In the last decade, numerous studies have linked language use to a wide range of psychological correlates~\cite{park2015automatic,ireland2010language}.
Reliable correlations of writing style (e.g., frequency of word use) with personality were found by some of the earliest works~\cite{pennebaker1999linguistic}. For example,
individuals scoring higher on extraversion used more positive emotion words (e.g., great, amazing, happy) whereas those higher in neuroticism were found to use first-person singulars (e.g., I, mine, me) more frequently. Personality detection using these type of differences in the use of language is referred to as the \textbf{closed vocabulary} approach.
\par
One of the most common closed-vocabulary methods for personality detection from text (especially social media) is LIWC. It categorizes the words into various psychologically relevant buckets like `function words'(e.g., articles, conjunctions, pronouns), `affective processes' (e.g., happy, nervous,  cried) and `social processes' (e.g., mate, talk, friend). It then counts the frequency of words in each of the buckets (over 60 psychologically relevant buckets) and then predicts the personality of the writer of the text. As this approach starts with predefined categories of words, it has been described as `closed-vocabulary' and most of the features extracted by LIWC are language dependent.
\par
Receptiviti API~\cite{golbeck2016predicting} is a tool built on LIWC which used for personality analysis from text. A user's personality prediction is made using psycholinguistic features. The Receptiviti API allows users to submit a text sample which is analyzed and outputs a graphical representation of the predicted traits which can also be viewed on a web-based interface. However, this API does not perform well on social media derived text as closed vocabulary methods have lower accuracies while dealing with micro-text in general.
\par
Personality traits may even be predicted from different sources of interpersonal communication (e.g., WhatsApp) on a mobile device.
\cite{saez2014system} have built an application which is in charge of compiling and sending information about the user to a server application, which then runs the classification algorithms. One line of research this paper follows is checking whether personality is linked to happiness of a person, however it didn't result in anything conclusive. This is one of the few research papers which uses the PEN theory for classification instead of the Big-Five. Standard machine learning algorithms such as J48, Random Forrest and SVM were tested.
\par
In the recent years, personality detection from social media, especially Twitter sentiment analysis has gained a lot of popularity. This might be partially due to the fact that Twitter data collection is direct and easily accessible through the Twitter API. \cite{kalghatgi2015neural} have used a combination of social behavior of the user (average number of links, average number of hash tags, average number of mentions, etc.) and grammatical information (average length of text, average number of positive and negative words, average number of special characters like comma, question mark, etc.) for user personality detection.
Using these attributes, a feature vector is constructed which is then fed into a multi-layer perceptron (MLP).
The novelty of this approach lies in running the whole set up on the Hadoop framework (Hadoop Distributed File System and Map Reduce) which enables personality prediction of N users concurrently.

\begin{figure}[ht]
\centering
\includegraphics[width=0.5\textwidth]{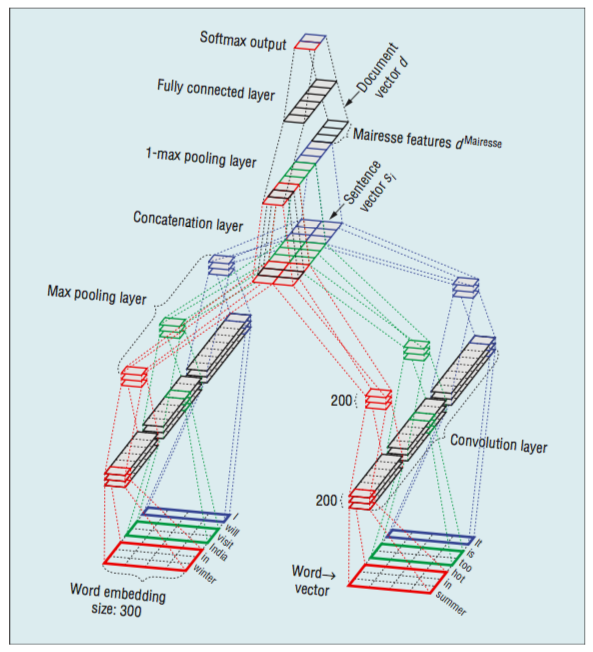}
\caption{\label{fig:text} CNN framework used by~\cite{majumder2017deep} makes use of Word2Vec embeddings for personality detection from text.}
\end{figure}

Apart from social media data, essays are also a popular mode of text and can be used for author profiling. A popular dataset is the James Pennebaker and Laura King's stream-of-consciousness essay dataset~\cite{pennebaker1999linguistic}. It contains 2,468 anonymous essays tagged with the authors' personality based on the Big-Five traits. \cite{majumder2017deep} makes use of a deep CNN for document level personality detection tasks.
The CNN helps in extracting the monogram, bigram and trigram features from the text and the architecture of this model is shown in Fig.\ref{fig:text}.
It is observed that the removal of neutral sentences gives a marked improvement in the prediction accuracy. Each word is represented in the input as a fixed-length feature vector using Word2Vec~\cite{mikolov2013efficient}, and sentences are represented as a variable number of word vectors.
In the end, document level Mairesse features (LIWC, MRC, etc.), totally 84 features, are concatenated with the feature vector extracted from the deep CNN. Lastly, this concatenated vector is then fed into a fully connected layer for the final personality trait predictions.
\cite{hernandezpredicting} have tried modeling temporal dependencies amongst sentences by feeding the input to Recurrent Neural Networks (RNNs). LSTMs were found to give better results compared to vanilla RNN, GRU, bi-LSTM. GloVe~\cite{pennington2014glove} word embeddings of 50D were used for predicting MBTI personality of people based on 50 posts on a particular social personality discussion website called PersonalityCafe.

\begin{table*}[t]
\centering
\caption{Popular datasets, divided based on mode of input.}
\begin{tabular}{@{}cccl@{}}
\toprule
\rowcolor[HTML]{EFEFEF} 
\textbf{Modality} & \textbf{Dataset}                                                                                                                                            & \textbf{\begin{tabular}[c]{@{}c@{}}Personality \\ Measure\end{tabular}} & \multicolumn{1}{c}{\cellcolor[HTML]{EFEFEF}\textbf{Description}}                                                                                                                                                                                                                           \\ \midrule
\rowcolor[HTML]{ECF4FF} 
Text              & Essays I~\cite{pennebaker1999linguistic}                                                                                                   & Big-Five                                                                & \begin{tabular}[c]{@{}l@{}}2,468 anonymous essays tagged with the author's personality traits. Stream-of-consciousness \\ essays written by volunteers in a controlled environment and the authors of the essays were \\ asked to label their own Big-Five personality traits\end{tabular} \\
\rowcolor[HTML]{ECF4FF} 
                  & Essays II~\cite{tausczik2010psychological}                                                                                                 & Big-Five                                                                & \begin{tabular}[c]{@{}l@{}}2,400 essays were labelled manually with personality scores for five different personality traits. \\ The data were then tuned and the regression scores were converted into class labels of five \\ different traits\end{tabular}                               \\
\rowcolor[HTML]{ECF4FF} 
                  & MBTI Kaggle\footnote{https://www.kaggle.com/datasnaek/mbti-type/}                                                                          & MBTI                                                                    & \begin{tabular}[c]{@{}l@{}}Contains the MBTI personality labels for over 8,600 people, along with 50 of their posts on the \\ myPersonality cafe forum\end{tabular}                                                                                                                        \\
\rowcolor[HTML]{ECF4FF} 
                  & MyPersonality\footnote{https://sites.google.com/michalkosinski.com/mypersonality}                                                         & Big-Five                                                                & \begin{tabular}[c]{@{}l@{}}myPersonality was a Facebook App that allowed its users to participate in psychological research\\ by filling in a personality questionnaire. However, from 2018 onwards they decided to stop sharing\\ the data with other scholars.\end{tabular}              \\
\rowcolor[HTML]{ECF4FF} 
                  & \begin{tabular}[c]{@{}c@{}}Italian \\ FriendsFeed~\cite{celli2010social}\end{tabular}                                                      & Big-Five                                                                & \begin{tabular}[c]{@{}l@{}}Sample of 748 Italian FriendFeed users (1,065 posts). The dataset has been collected from \\ FriendFeed public URL, where new posts are publicly available\end{tabular}                                                                                         \\
\rowcolor[HTML]{FFFFC7} 
Audio             & \begin{tabular}[c]{@{}c@{}}AMI \\ Meeting Corpus~\cite{carletta2005ami}\end{tabular}                                                       & Big-Five                                                                & \begin{tabular}[c]{@{}l@{}}Video and audio recordings of monologues, dialogues and multi-party discussions with annotations \\ to perceived personality which is found by the BFI-10 questionnaire\end{tabular}                                                                             \\
\rowcolor[HTML]{FFFFC7} 
                  & Aurora2 corpus~\cite{hirsch2000aurora}                                                                                                     & -                                                                       & \begin{tabular}[c]{@{}l@{}}Connected digit corpus which contains 8,440 sentences of clean and multi-condition training data and \\ 70,070 sentences of clean and noisy test data\end{tabular}                                                                                              \\
\rowcolor[HTML]{FFFFC7} 
                  & \begin{tabular}[c]{@{}c@{}}CMU self-recorded \\ database~\cite{polzehl2010automatically}\end{tabular}                                      & Big-Five                                                                & \begin{tabular}[c]{@{}l@{}}Experimental database with a professional speaker was recorded and generated the Big-Five factor \\ scores for the recordings by conducting listening test using the NEO-FFI personality inventory.\end{tabular}                                                \\
\rowcolor[HTML]{FFFFC7} 
                  & \begin{tabular}[c]{@{}c@{}}Columbia \\ deception corpus~\cite{levitan2015cross}\end{tabular}                                               & -                                                                       & \begin{tabular}[c]{@{}l@{}}Balanced corpus includes data from 126 (previously unacquainted) subject pairs, constituting \\ 93.8 hours of speech in English\end{tabular}                                                                                                                    \\
\rowcolor[HTML]{FFCCC9} 
Visual            & PhychoFlikr dataset~\cite{cristani2013unveiling}                                                                                           & Big-Five                                                                & \begin{tabular}[c]{@{}l@{}}Top 200 images favorited by 300 Flikr users (hence totally 60,000 images) labelled with their \\ personality traits\end{tabular}                                                                                                                                \\
\rowcolor[HTML]{FFCCC9} 
                  & \begin{tabular}[c]{@{}c@{}}First Impressions V2 \\ (CVPR'17)\footnote{http://chalearnlap.cvc.uab.es/dataset/24/description/}\end{tabular} & Big-Five                                                                & \begin{tabular}[c]{@{}l@{}}The first impressions data set, comprises 10,000 clips (average duration 15s) extracted from more\\ than 3,000 different YouTube high-definition (HD) videos of people facing and speaking in English \\ to a camera\end{tabular}                               \\ \bottomrule
\end{tabular}
\label{table2}
\end{table*}

\par
It was found that combining commonsense knowledge with psycho-linguistic features resulted in a remarkable improvement in the accuracy~\cite{poria2013common}.
SenticNet~\cite{camnt5} is a popular tool used for extracting commonsense knowledge along with associated sentiment polarity and affective labels from text. It is one of the most useful resources for opinion mining and sentiment analysis.
These features are used as inputs to five Sequential Minimal Optimization based supervised classifiers for the five personality traits, which are predicted by each of the classifiers independently.
\par
While most researchers assume the five personality traits to be independent of each other, several studies~\cite{shaver1992attachment,judge1999big} claim that there exists certain correlations among the traits and it is inaccurate to build 5 completely independent classifiers.
\cite{zuo2013weighted} have modeled the inter-trait dependencies using a weighted ML-KNN algorithm which uses information entropy theory~\cite{jaynes1982rationale} to assign weights to the extracted features.
The linguistic and correspondingly dependent emotional features are extracted from text which can be then discretized on the basis of Kohonen's feature-maps algorithm~\cite{kohonen1990self}.
\\
Instead of relying on prior word or category judgments, \textbf{open-vocabulary} methods rely on extracting a comprehensive collection of language features from text. These methods characterize a sample text by the relative use of non-word symbols (e.g., punctuation, emoticons), single uncategorized words, multi-word phrases and clusters of semantically related words identified through unsupervised methods (topics). Typically, Latent Dirichlet Allocation (LDA)~\cite{blei2003latent} is used for forming clusters of semantically related words.
\par
MyPersonality dataset is one of the famous textual social media datasets which comprises of status updates of over 66,000 Facebook users. Each of these volunteers also completed the Big-Five personality test. \cite{park2015automatic} use this dataset for automatic personality detection from the language used in social media. Firstly, features are extracted using standard techniques for text. Then, the dimensionality of these features is reduced followed by Ridge Regression~\cite{hoerl1970ridge}. There is an individual regressor for prediction of each of the five personality traits. An end to end deep learning model was trained on a subset of this dataset (Facebook user status) where the authors~\cite{yu2017deep} experimented with n-grams (extracted with CNN) and bi-directional RNNs (forward and backward GRU). Instead of using the pre-trained Word2Vec model, they trained the word embedding matrix using the skip-gram method~\cite{mikolov2013efficient} to incorporate internet slang, emoticons and acronyms. To improve accuracy, they perform late fusion of non-lexical features extracted from text and the penultimate layer of the neural network. This concatenated vector is then fed into a softmax output layer.
\par
\cite{liu2016analyzing} have used deep learning based models in combination with the atomic features of text, the characters. An individual's personality traits are predicted using hierarchical, vector word and sentence representations. This work is based on~\cite{ling2015finding}, which provides a language independent model for personality detection. Instead of making use of standard word embeddings (e.g., GloVe or Word2Vec), the word vector representation is formed from a bi-RNN using GRU~\cite{chung2014empirical} as the recurrent unit. GRU is less computationally expensive than the LSTM, but results in similar performance.
These words are fed to another bi-RNN which gives the representation of the sentence, followed by a feed forward neural network for the prediction of the values of the five traits.
One of the best uses of this deep learning based character-level word encoding is a corpus of Tweets. The data are noisy and consist of a large number of micro-texts, hence this model performs much better than the conventional LIWC/BoW models.
\par
Recently, NLP tools have been developed for language-independent, unsupervised personality recognition from unlabeled text (PR2~\cite{celli2014pr2}). The PR2 system exploits language-independent features extracted from LIWC and MRC such as punctuation, question marks, quotes, exclamation marks,  numbers, etc.
Only feature values which are above average in the input text are mapped to personality traits.
It was observed that text containing more punctuation fires negative correlations with extraversion and openness to experience. This approach is not as accurate as the state of the art in the field, but it much more computationally efficient along with being independent of language.
\par
Instead of relying just on text as the only form of input, researchers at Nokia have developed a software~\cite{chittaranjan2011s} which make use of data from smart phones including information about which applications were used and how often, music preferences, anonymous call logs, SMS logs and Bluetooth scans for personality detection. The relationship between personality traits and usage features was systematically analyzed using standard correlation and multiple regression analyses. SVM and C4.5 classifiers were used to classify users' personality traits.
\par
Social media is a huge source of people's opinions and thoughts  and analyzing it can give important insights into people's personality. \cite{gonzalez2015tweets} have used stylistic features which can be divided in two groups: character N-grams and POS N-grams. The grammatical sequence of the writer is obtained from POS tagging by making use of the open source tool, Freeling. Character N-grams and POS N-grams are extracted from the POS and class files to create a feature vector which is then fed into a linerSVM for personality detection. 

\cite{sun2018personality} have introduced abstract feature combination based on closely connected sentences which they call the Latent Sentence Group. The authors use bidirectional LSTMs, concatenated with CNN to detect user's personality using the structures of text. 
\par
Sometimes, the input text may be a collection of chats between two individuals.
Detecting personality from conversations is a harder task than with simple text as one needs to accurately model temporal dependencies. Interactions in dyadic conversations have various degrees of mutual influence caused by turn-taking dialogues between two individuals. Recurrent networks are good for taking care of language dependencies. \cite{su2016exploiting} make use of vanilla RNNs for modeling short term temporal evolution in conversation. A 255-dimensional linguistic feature vector is coupled Hidden Markov Model (C-HMM) is employed for detecting the personalities of two speakers across speaker turns in each dialog by using long-term turn-taking temporal evolution and cross-speaker contextual information. The Mandarin Conversational Dialog Corpus~\cite{tseng2004processing} was used as the dyadic conversation corpus for training and evaluation of the model.
In many studies, how the personality `appears' is approximated rather than the `true' personality (after taking the personality questionnaire) of an individual. Sometimes, trained judges label an individual's perceived personality depending on their behavior, giving us an insight into how they are perceived by those around them.


\subsection{\textbf{Audio}}
We observed that majority of the recent models for personality detection from audio modality work in two separate stages, feature extraction followed by feeding the features to a classifier. The audio descriptors can be sub-divided into broadly 7 groups, intensity, pitch, loudness, formants, spectrals, MFCC and other features. Some researchers use one or more of these feature groups to assess personality which is derived at the utterance level~\cite{polzehl2010automatically}.
These audio features  (e.g., intensity, pitch, MFCC) are extracted using Praat acoustic analysis program and fed to a SVM classifier.
The authors of this paper~\cite{polzehl2010automatically} found that the five personality traits are interdependent and the variation in one of the traits results in observable changes in the other four. The dataset is created by recording the utterances and labelling them by having a trained speaker mimic a particular type of personality. Hence, the accuracy of the training and evaluation sets is solely dependent on how good the trained speaker is at mimicking the voice of a certain personality.
\par
Some researchers claim that non-linguistic features (prosody, overlaps, interruptions and speech activity) outperform linguistic features (dialog acts and word n-grams) for perceived personality detection. \cite{valente2012annotation} have tested their model on spoken conversations from the AMI corpus dataset~\cite{carletta2007unleashing}, which is a collection of meetings captured in specially instrumented meeting rooms along with the audio and video of each of the participants. The novelty of their approach lies in incorporating dialog act tags (which capture the intention of the speaker in the discussion) as one of the input features fed into the classifier.
Each speaker was been labeled with 14 Dialog Act tags and a correlation with their perceived personality traits was observed (especially extraversion).
Speech activity features (the total and relative amount of speech time per speaker, average duration of pauses per speaker, etc.) and sentence features (total number of sentences, average duration, maximum duration, etc.)
are also included in the algorithm using a Booster algorithm with multi-class Boosting~\cite{schapire2000boostexter}.
\par
Recently, end-to-end approaches which involve training deep neural networks have gained wide spread popularity. They yield the state-of-the-art accuracies along with automatically extracting the necessary acoustic features. \cite{palaz2015analysis} analyze the CNN to understand the speech information that is modeled between the first two convolution layers. The first convolutional layer acts as the filter bank and automatically learns to extract the relevant features from the unprocessed audio wave input.
The key difference in this paper is that temporal raw speech is fed directly to the CNN instead of the conventional two step procedure of first extracting the feature vectors (e.g., MFCC) followed by feeding to a classifier.
\par
It is often observed that knowing the personality traits of an individual may in turn help in other tasks, such as deception detection.
\cite{levitan2016identifying} have used AdaBoost~\cite{freund1996experiments} and Random forest for personality detection, which is then used as input to their deception detection algorithm. They have made use of acoustic, prosodic and linguistic features (LIWC) for personality prediction. The contribution of each of the LIWC features is then analyzed. It was found that for NEU-score, 'money' and 'power' dimensions are the most useful, for EXT-score, 'focusfuture' and 'drives', for CON-score, 'time' and 'work' are the highest. Most of these are intuitive and show the power of using LIWC features for personality detection. The model performed better in deception detection after incorporating the personality traits of the subject and was evaluated on the Columbia X-Cultural Deception Corpus.

\subsection{\textbf{Visual}}

Physiognomy~\cite{kamenskaya2008recognition} is the art of determining the character or personality traits of an individual from the features of the body, especially the face. Researchers have found that the face provides most of the descriptive information for perceived personality trait inference~\cite{bruce1986understanding,willis2006first}.
A non deep learning approach which involves the analysis of facial features for personality prediction is given by~\cite{kamenskaya2008recognition}. Their focus lies on extracting physical features from images like shape of nose, body shape (fat, muscular, thin), shape of eyebrows and finding their correlation with personality.
Sometimes people of similar personality tend to have similar preferences.
\cite{cristani2013unveiling} have explored this idea and have studied if there is a relation between the aesthetic preferences of a person (e.g., pictures liked by them) and their personality. Features extracted from the images could be divided into two types, aesthetic (colorfulness, use of light, GLCM-features, etc.) and content (faces, objects, GIST descriptors), which are then fed into a lasso regressor.

\cite{liu2016analyzing} have predicted personality traits of a person from the choice of their Twitter profile picture. The model was trained on data from more than 66,000 users whose personality type had been predicted from what they had tweeted previously, using the state-of-the-art techniques from text modality. A correlation was found between the aesthetic and facial features of the profile picture (extracted using face++ and EmoVu API) and users' personality traits. Agreeableness and Conscientiousness users display more positive emotions in their profile pictures, while users high in openness prefer more aesthetic photos. Self-presentation characteristics of the user are captured by the 3D face posture, which includes the pitch, roll and yaw angle of the face and eye openness (using EmoVu API). Appealing images tend to have increased contrast, sharpness, saturation and less blur, which is the case for people high in Openness.

\begin{figure}[h]
\centering
\includegraphics[width=\linewidth]{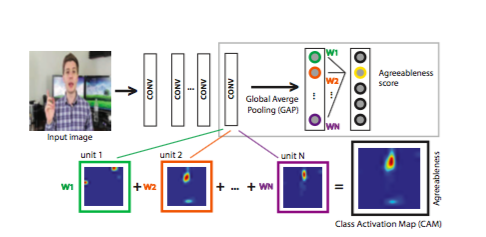}
\caption{\label{fig:visual} Class activation map used by~\cite{ventura2017interpreting} for interpreting CNN models.}
\end{figure}

As in many other computer vision problems, deep CNNs have performed remarkably well in the task of personality detection from images.
However, despite of the success in terms of performance, CNN is widely regarded as the magic black box. It is not known what are the various internal representation which emerge in various hidden layers of the neural network. \cite{ventura2017interpreting} presents an in-depth study on understanding why CNN models perform surprisingly well in this complex problem.
They make use of class activation maps for visualization and it is observed that activation is always centered around the face as shown in Fig.\ref{fig:visual}. The neural network specifically focuses on the facial region, mainly areas corresponding to the eyes and mouth to discriminate among different personality traits and predict their values.

The deeper we go in the CNN, the more the layers are specialized in identifying high level features such as the eyes, nose, eyebrows, etc. Several Action Units (AU) such as raised eyebrows, dimple, etc. (subset of the FACS) are taken and a good correlation was observed between AU and perceived personality. A feature vector of the AUs was fed directly into a simple linear classifier and accuracy close to the state of the art was achieved.

\begin{table*}[t]
\centering
\caption{Brief description of the important personality-detection models.}
\begin{tabular}{@{}ccl@{}}
\toprule
\rowcolor[HTML]{EFEFEF} 
\textbf{Modality}                            & \textbf{Paper}                                    & \multicolumn{1}{c}{\cellcolor[HTML]{EFEFEF}\textbf{Architecture}}                                                                                                                                              \\ \midrule
\rowcolor[HTML]{ECF4FF} 
Text                                         &~\cite{majumder2017deep}          & 1D convolutions to extract n-grams combined with Mairesse features                                                                                                                                             \\
\rowcolor[HTML]{ECF4FF} 
                                             &~\cite{poria2013common}           & \begin{tabular}[c]{@{}l@{}}Combined LIWC with MRC features extracted from text along with commonsense knowledge using\\ sentic computing techniques\end{tabular}                                              \\
\rowcolor[HTML]{ECF4FF} 
                                             &~\cite{hernandezpredicting}       & GloVe embedding fed into an deep RNN (LSTM)                                                                                                                                                                    \\
\rowcolor[HTML]{ECF4FF} 
                                             &~\cite{celli2012unsupervised}     & \begin{tabular}[c]{@{}l@{}}Used a subset of Mairesse features along with unsupervised learning methods to measure correlation\\ between features and personality traits\end{tabular}                           \\
\rowcolor[HTML]{FFFFC7} 
Audio                                        &~\cite{valente2012annotation}     & \begin{tabular}[c]{@{}l@{}}Audio features such as prosodic features, speech activity features, word n-grams and dialogue act \\ tags were extracted and fed into simple ML classifiers\end{tabular}            \\
\rowcolor[HTML]{FFFFC7} 
                                             &~\cite{polzehl2010automatically}  & \begin{tabular}[c]{@{}l@{}}Cepstral features of speech such as MFCC, Zero Crossing Rate (ZCR), intensity, pitch, loudness,\\formants were fed into a SVM regressor\end{tabular}                              \\
\rowcolor[HTML]{FFFFC7} 
                                             &~\cite{levitan2016identifying}    & \begin{tabular}[c]{@{}l@{}}Combination of LIWC and prosodic features fed into ML classifiers such as AdaBoost \\ and RandomForest\end{tabular}                                                                 \\
\rowcolor[HTML]{FFCE93} 
Visual                                       &~\cite{rojas2011automatic}        & \begin{tabular}[c]{@{}l@{}}Histogram of Oriented Gradients (HOG) for modeling the face, EigenFace and specific \\ points on the face were taken and fed into SVM, Gradient Boosting, bTree\end{tabular}        \\
\rowcolor[HTML]{FFCE93} 
                                             &~\cite{biel2012facetube}          & \begin{tabular}[c]{@{}l@{}}CERT used for extracting facial expressions \\ followed by thresholding as well as Hidden Markov Models\end{tabular}                       \\
\rowcolor[HTML]{FFCE93} 
\multicolumn{1}{l}{\cellcolor[HTML]{FFCE93}} &~\cite{gurpinar2016combining}     & \begin{tabular}[c]{@{}l@{}}Transfer Learning, pre-trained VGG-Face and VGG-19 to extract facial as well as background \\ features followed by regularized regression with a kernel ELM classifier\end{tabular} \\
\rowcolor[HTML]{FFCE93} 
\multicolumn{1}{l}{\cellcolor[HTML]{FFCE93}} &~\cite{al2014face}                & Polynomial and Radial Basis Function (RBF) kernels for SVMs                                                                                                                            \\
\rowcolor[HTML]{FFCCC9} 
Multimodal                                   &~\cite{guccluturk2017visualizing} & \begin{tabular}[c]{@{}l@{}}Facial features extracted by VGG-Face, Scene features by VGG-VD19 and combined with audio \\ features using Random Forest based score level fusion\end{tabular}                     \\
\rowcolor[HTML]{FFCCC9} 
\multicolumn{1}{l}{\cellcolor[HTML]{FFCCC9}} &~\cite{kindiroglu2017multi}       & Multi task learning of learning leadership and extraversion simultaneously                                                                                                                                     \\ \bottomrule
\end{tabular}
\label{table3}
\end{table*}

\begin{figure}[h]
\centering
\includegraphics[width=0.5\textwidth]{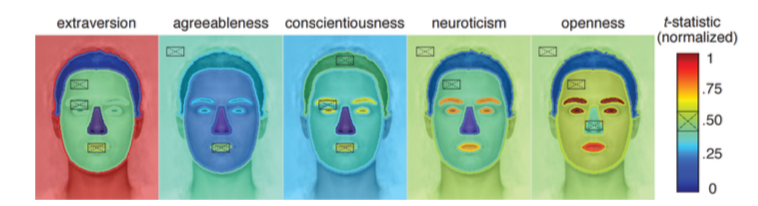}
\caption{\label{fig:visual2} Segment-level occlusion analysis. Each image shows the changes in the prediction of the corresponding trait as a function of a predefined region~\cite{guccluturk2017visualizing}.}
\end{figure}

The contribution of different parts of the image on personality trait prediction was studied by~\cite{guccluturk2017visualizing}  using segment level occlusion analysis. Video frames were segmented into the following six regions with a deep neural network: background region, hair region, skin region, eye region, nose region and mouth region. It was observed that each region modulates at least one trait, and different traits were modulated by different regions. Background, skin and mouth regions modulated the fewest traits. Occlusion of background region increased extraversion trait but decreased conscientiousness trait. Occlusion of skin region decreased agreeableness and conscientiousness traits. Occlusion of mouth region increased neuroticism and openness traits. Finally, the eye region negatively modulates agreeableness trait, and positively modulates neuroticism and openness traits. Both the face region and the background region which includes clothing, posture and body parts have been shown to contain information regarding the perceived personality traits as can be seen in Fig.\ref{fig:visual2}.

Psychological studies suggest that humans infer trait judgments of another person in just a few milliseconds after meeting them, that is from their first impression.
First impressions and perceived personality are in-fact quite similar.
One of the most important and widely used dataset in the field of multimodal perceived personality recognition is the ChaLearn First Impressions dataset. It comprises of 10,000 clips (average duration 15s) extracted from more than 3,000 different YouTube high-definition (HD) videos of people facing and speaking into the camera. \cite{chen2016overcoming} have proposed several methods for eliminating worker bias in labelling large datasets and have tested their method on the ChaLearn dataset. \cite{gurpinar2016combining} make use of the visual modality from this dataset for testing their model, which combines facial and ambient features extracted by a pre-trained deep CNN. These features are then fed into a kernel Extreme Learning Machine Regressor~\cite{huang2012extreme}. The main contribution of this work is the combination of facial and ambient features along with making effective use of transfer learning. 
\par
A famous algorithm for detecting faces in images is the Viola Jones algorithm. \cite{al2014face} have used this algorithm in the preprocessing stages after converting the image to grayscale, histogram equalizations (increases contrast) and rescaling. Then, EigenFace~\cite{turk1991eigenfaces} features are extracted and inputted to a RBF SVM trained on Sequential Minimal Optimization approach. This approach was tested on a database created by making judges independently label images in the FERET corpus with the perceived personality traits. It is assumed that the judges have no ethnic or racial bias.
\par
\cite{rojas2011automatic} have studied if there are specific structural relations among facial points that predict perception of facial traits. They have made a comparison between two approaches, first which uses the whole appearance and texture of the face (holistic approach)~\cite{roberto1993face} or just locations of specific fiducial facial points (structural approach). The holistic approach captures facial appearance information in two ways. Firstly, via the EigenFaces~\cite{turk1991eigenfaces} method which analyzes the pixel information or secondly, making use of Histogram of Oriented Gradients (HOG), which captures facial appearance and shape. These extracted features are then fed into standard ML classifiers such as SVM or BTrees. Instead of using the Big-Five model, this paper predicts 9 different personality traits such as trustworthy, dominant, extroverted, threatening, competence, likability, mean, attractive and frightening. It was observed that HOG model found more correlation between the facial features and personality traits as compared to EigenFace.
\par
A Vlog refers to a blog in which the postings are primarily in video form. A correlation between the facial expressions and personality of vloggers was analyzed by~\cite{biel2012facetube}. In this paper, Computer Expression Recognition Toolbox (CERT) was used to determine the facial expressions of vloggers. CERT predicts 7 universal facial expressions (joy, disgust, surprise, neutral, sadness, contempt and fear). They found that facial expressions of emotions with positive valence such as `joy' and `smile' showed almost exclusively positive effects with personality impressions. For example, `smile' is correlated with extraversion and agreeableness, among others.

\par
Although the aforementioned approaches yield good results, they are very computationally expensive. \cite{eddine2017personality} use a temporal face texture based approach for analyzing facial videos for job screening, which is much more computationally economical.
After the preprocessing step, which involves face identification, 2D pose correction and cropping the region of interest (ROI), texture features of the image are extracted using Local Phase Quantization (LPQ) and Binarized Statistical Image features (BSIF).
These features are fed to a Pyramid Multi Layer followed by five nonlinear Support Vector Regressors, one for each of the Big-Five personality traits. Finally, this perceived personality prediction is then input to a Gaussian Process Regression (GPR) which predicts a binary value corresponding to whether the candidate should be called for a job interview or not.

\subsection{\textbf{Bimodal}}

Most of the multimodal architectures for personality trait recognition are bimodular with the fusion of features from the audio and visual modalities.
\cite{kindiroglu2017multi} make use of multi task learning for recognizing extraversion and leadership traits from audio and visual modalities extracted from meeting videos of the Emergent Leader (ELEA) corpus. Predicting extraversion trait has attracted the attention of psychologists because it is able to explain a wide range of behaviors, predict performance and also assess the risk of an individual. The network architecture is initially trained on the larger VLOG corpus and afterwards on the ELEA corpus, hence effectively making use of transfer learning. Feature selection techniques such as Maximum Relevance Minimum Redundancy (MRMR) filtering feature selection results in a slight increase (about 2\%) in performance.
\par
Visualizing Deep Residual Networks (DRN) used for perceived personality analysis is done by~\cite{guccluturk2017visualizing}  with the focus on explaining the workings of this end-to-end deep architecture. This DRN is a 2 stream network with one stream for the visual modality and the other one for the audio modality.
What information do perceived personality trait recognition models rely on to make predictions?
\cite{guccluturk2017visualizing} concluded that female faces with high levels of all traits seemed to be more colorful with higher contrast compared to those with low levels, who were more uniformly colored with lower contrast. Furthermore, looking at the unisex average faces, a bias for female faces for the high levels and male faces for the low levels was observed, more so in the case of average faces based on annotations.
For instance, high levels of all traits were more bright and colorful, and with more positive expressions in line with the results observed by~\cite{walker2016changing}.

\begin{figure}[ht]
\centering
\includegraphics[width=0.5\textwidth]{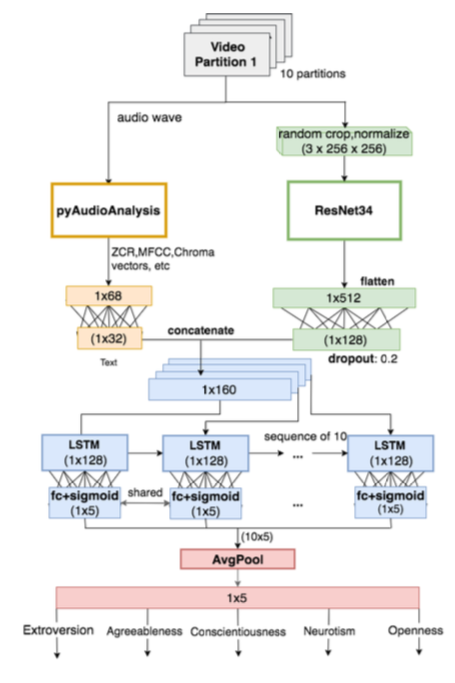}
\caption{\label{fig:multimodal} Deep Bimodal LSTM Model Architecture used by~\cite{Stanford2017PredictionOP}.}
\end{figure}

The Deep Bimodal Regression (DBR)~\cite{zhang2016deep} framework achieved the highest accuracy in the ChaLearn Challenge 2016 for perceived personality analysis. DBR consists of three main parts: the visual modality regression, audio modality regression and the last part is the ensemble process for fusing information of the two modalities. The network is trained end-to-end without the usage of any feature engine. The traditional CNN architecture is modified by discarding the fully connected layers. Also, the deep descriptors of the last convolution layer are both averaged and max pooled, followed by concatenation into a single 1024D vector (512 values from max pool and the other 512 values from average pooling). Finally, a regression (fc+sigmoid) layer is added for end-to-end training. This modified CNN model is called Descriptor Aggregation Network (DAN). For the audio modality, the log filter bank (logfbank)~\cite{davis1990comparison} features are extracted from the original audio of each video. Based on the logfbank features, a linear regressor is trained to obtain the Big-Five trait predictions. Finally, the two modalities are lately fused by averaging the five predicted scores of the visual and audio model.
\par
\cite{rai2016bi} propose a much more computationally efficient bimodal regression framework for the same, however with a slight decrease in accuracy. Instead of a single 15s audio clip, multiple audio segments (2-3 seconds length) are extracted from each video which increases the number of training samples resulting in considerably better results. FFMPEG is used for extracting audio clips from the original videos, followed by the OpenSMILE framework for extracting audio features. There are many studies~\cite{shaver1992attachment,judge1999big} which claim that the Big-Five personality traits are not independent of each other. The strength of a particular trait is quantified by the combination of two components, i.e., a global component and a trait specific component. In this model, each trait specific model is learned using a global component along with a trait specific component. Six models are created to predict the five personality traits where one model represents the global component and the other five represent the trait-specific components.

\cite{An2018} have combined audio and text modalities for personality detection. The authors use acoustic-prosodic low level
descriptor (LLD) features, word
category features from LIWC and word scores for pleasantness, activation and imagery from the Dictionary of Affect in Language (DAL)~\cite{whissell1986dictionary}. A concatenation of these features are then fed into a MLP.
They have compared this with a network consisting of word embedding with LSTM, and showed that the bimodal MLP network performs best on the myPersonality corpus. The MLP generalizes better when faced with out-of-vocabulary words. This points to the promise of acoustic-prosodic features, which are more robust with respect to language. They also find that models with early and late fusion of features achieve similar performance.
\par
\cite{subramaniam2016bi} makes use of temporally ordered deep audio and stochastic visual features for bimodal first impressions recognition. The novel idea of this paper is predicting perceived personality traits by making use of volumetric 3D convolution~\cite{ji20133d} based deep neural network. The temporal patterns in the audio and visual features are learned using a LSTM based deep RNN. Both the models concatenate the features extracted from audio and visual data in a later stage and finally feed into a fully connected layer.
\par
\cite{Stanford2017PredictionOP} follows a similar approach in trying to model the temporal dependencies by making use of LSTM. Audio features such as ZCR, MFCC and chroma vectors are extracted using PyAudio Analysis, whereas images are fed to a ResNet34~\cite{he2016deep} architecture after random cropping and normalization. This architecture is shown in Fig.\ref{fig:multimodal} and achieves state-of-the-art performance for conscientiousness and openness traits. \cite{gurpinar2016multimodal} achieve state-of-the-art performance in this field by feeding the extracted feature vectors to a Extreme Learning Machine (ELM) and then fusing their predictions.

Acoustic features are extracted via the popularly used openSMILE tool while pre-trained Deep Convolutional Neural Network, VGG-VD-19~\cite{simonyan2014very} was used to extract facial emotion and ambient information from images. Local Gabor Binary Patterns from Three Orthogonal Planes (LGBP-TOP)~\cite{zhang2005local} along with VGGFace was used for facial feature extraction. The best results are achieved with multi-level fusion of one acoustic and three visual sub-systems which are fed into an ELM and then combined with score level fusion.

Apart from perceived personality analysis of the Big-Five traits, an individual's attitudes such as amusement, impatience, friendliness, enthusiasm, frustration can also be estimated using similar architectures.
\cite{madzlan2014automatic} automatically recognizes attitudes of Vloggers using prosodic and visual feature analysis. Prosodic features of audio such as pitch, intensity, quality, duration of utterance, etc. were extracted using TCL/TK script. Prosodic analysis of the attitudes shows 'impatience' having the highest pitch whereas 'frustration' the lowest. However, the analysis of visual modality is simply done by correlating the absolute movement of landmarks (eyebrows, mouth, etc.) of the face with respect to attitude.

\cite{zhao2019personalized} is one of the only papers to computationally study the influence of personality traits on emotion. The authors have jointly modeled physiological signals with personality. They have presented Vertex-weighted Multimodal Multitask Hypergraph Learning (VM2HL) learning model, where vertices are (subject, stimuli) pairs. Hyperedges formulate the relationship between personality and physiological signals.
Hyperedges, modalities and vertices are given different levels of importance by introducing weights which are learned during training. 

\subsection{\textbf{Trimodal}}
While most of the work focuses on fusion of inputs from audio and visual sources, \cite{gorbova2017automated} creates an additional stream that incorporates input from the text modality. These personality predictions are then used for automated screening of job candidates. They have achieved comparable performance to state-of-the-art while making use of a fully connected neural network with only 2 hidden layers.
Facial features are extracted using OpenFace~\cite{baltruvsaitis2016openface}, an open source facial feature extractor toolkit. Audio features such as MFCC, ZCR, spectral energy distribution and speaking rate are extracted using OpenSMILE~\cite{eyben2010opensmile} and SenticNet is used for polarity detection from text.

\begin{figure}[h]
\centering
\includegraphics[width=0.45\textwidth]{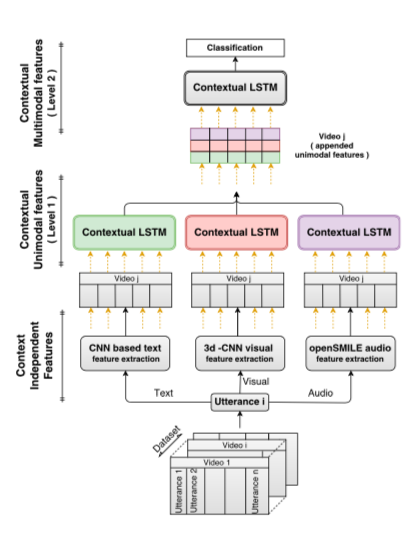}
\caption{\label{fig: trimodal} Hierarchical architecture for extracting context- dependent multimodal utterance features~\cite{poria2017context}.}
\end{figure}

\cite{poria2017context} carries out multimodal sentiment analysis and make use of LSTMs to capture the context along with dependencies in between utterances. Initially, context independent features are extracted from each of the modalities. Features from the visual modality are extracted using 3D-CNN, OpenSMILE is used to capture low level descriptors from audio and Word2Vec embeddings followed by CNN for the text modality. To capture the flow of informational triggers across utterances, a LSTM-based RNN is used in a hierarchical scheme. Lastly, this is then fed to a dense layer for the final output. This architecture performs better than the state of the art on IEMOCAP~\cite{busso2008iemocap}, MOUD~\cite{perez2013utterance} and MOSI~\cite{zadeh2016mosi} datasets and is shown in Fig.\ref{fig: trimodal}.

Deep networks being trained on multimodal data are generally very computationally intensive. \cite{vo2018multimodal} have proposed a multimodal neural network architecture that combines different type of input data of various sizes with Discriminant Correlation Analysis (DCA)~\cite{haghighat2016discriminant} Feature Fusion layer with low computational complexity.
The fused features are used as inputs for layers in the Mixture Density Network to adjust for the information loss due to feature fusion. The final prediction is made using a state-of-art cascade network built on advanced gradient boosting algorithms.
\par

Instead of having different trained neural networks for each of the
5 personality traits, \cite{kampman2018investigating} train a single neural network with a multiclass approach with 3 channels (one for each of the modalities). Features are extracted from each of the 3 modalities using CNNs. The authors have experimented with using early fusion (concatenation of vectors to obtain shared representations of the input data) along with late fusion (each of the modalities makes an individual prediction and the prediction of the network is some weighted average of these predictions).

\section{Results and Discussions}
\label{Results and Discussions}

\begin{table*}[t]
\centering
\caption{Performance of the state-of-the-art methods on popular
  personality-detection datasets.}
\begin{tabular}{@{}clccc@{}}
\toprule
\rowcolor[HTML]{EFEFEF} 
\textbf{Modality}                            & \multicolumn{1}{c}{\cellcolor[HTML]{EFEFEF}\textbf{Dataset}}                 & \textbf{Personality Measure} & \textbf{Mean Best Accuracy} & \textbf{Paper}                                    \\ \midrule
\rowcolor[HTML]{ECF4FF} 
Text                                         & Essays I~\cite{pennebaker1999linguistic}                    & Big 5                        & 57.99                       &~\cite{majumder2017deep}          \\
\rowcolor[HTML]{ECF4FF} 
                                             & Essays II~\cite{tausczik2010psychological}                  & Big 5                        & 63.6                        &~\cite{poria2013common}           \\
\rowcolor[HTML]{ECF4FF} 
                                             & MBTI Kaggle                                                                  & MBTI                         & 67.77                       &~\cite{hernandezpredicting}       \\
\rowcolor[HTML]{ECF4FF} 
                                             & Italian FriendsFeed~\cite{celli2010social}                  & Big 5                        & 63.1                        &~\cite{celli2012unsupervised}     \\
\rowcolor[HTML]{FFFFC7} 
Audio                                        & The AMI Meeting Corpus~\cite{carletta2005ami}               & Big 5                        & 64.84                       &~\cite{valente2012annotation}     \\
\rowcolor[HTML]{FFFFC7} 
                                             & CMU self recorded database~\cite{polzehl2010automatically}  & Big 5                        & -                           &~\cite{polzehl2010automatically}  \\
\rowcolor[HTML]{FFFFC7} 
                                             & Columbia deception corpus~\cite{levitan2015cross}           & -                            & -                           &~\cite{levitan2016identifying}    \\
\rowcolor[HTML]{FFCE93} 
Visual                                       & Face Evaluation~\cite{oosterhof2008functional}              & -                            & -                           &~\cite{rojas2011automatic}        \\
\rowcolor[HTML]{FFCE93} 
                                             & YouTube Vlogs                                                                & Big 5                        & 0.092 (R-squared)           &~\cite{biel2012facetube}          \\
\rowcolor[HTML]{FFCE93} 
\textit{\textbf{}}                           & ChaLearn First Impressions Dataset~\cite{ponce2016chalearn} & Big 5                        & 90.94                       &~\cite{gurpinar2016combining}     \\
\rowcolor[HTML]{FFCE93} 
\textit{\textbf{}}                           & Color FERET database~\cite{phillips1998feret}               & Big 5                        & 65                          &~\cite{al2014face}                \\
\rowcolor[HTML]{FFCCC9} 
Multimodal                                   & ChaLearn First Impressions Dataset~\cite{ponce2016chalearn} & Big 5                        & 91.7                        &~\cite{guccluturk2017visualizing} \\
\rowcolor[HTML]{FFCCC9} 
\multicolumn{1}{l}{\cellcolor[HTML]{FFCCC9}} & ELEA (Emergent LEAder) corpus~\cite{sanchez2011audio}       & Leadership / Extraversion    & 81.3                        &~\cite{kindiroglu2017multi}       \\ \bottomrule
\end{tabular}
\label{table4}
\end{table*}

Table~\ref{table4} provides a systematic overview of the classification performance on popular personality-detection datasets.
There has been a large increase in the number of works published in the field of personality detection in the last couple of years. This is due to the huge number of industrial applications of personality detection, as discussed in Section~\ref{Applications}. Recently, systematically annotated datasets have been made publicly available and, hence, research can now be focused on creating better models and architectures rather than on data procurement and preprocessing. The CVPR First Impressions dataset~\cite{ChalearnDataset} is the most popular multimodal dataset in personality detection. We see that there are many datasets for the Big-Five personality measure, however other personality measures (e.g., MBTI, PEN, 16PF) are lacking resources. The reliability of personality tests comes into question as it sometimes happens that when a person takes the test it gives a certain personality, however when the same person takes the same test again, it gives a different type of personality. \cite{escalante2018explaining,escalante2017design} study the interpretability and explainability of deep learning models in the context of computer vision for looking at people tasks.
The MBTI personality measure is the most popular personality measure used across the world right now. \cite{furnham1990can} have examined the relationship between subjects' actual test derived scores and their estimates of what those scores would be. Results showed significant positive correlations in majority of the dimensions.
MBTI traits are harder and more complex to predict than the Big-Five traits~\cite{furnham1996big}, but we see current deep learning models are starting to achieve good accuracy in this area as well.

We see that the state of the art in personality detection has been achieved using deep learning techniques along with multimodal fusion of features. Techniques involving bimodal fusion are currently very popular. One can see that there is lot of scope for research and exploration for trimodal fusion for personality detection as presently, relatively few works have explored the fusion of all 3 modalities. If an individual's personality could be predicted with a little more reliability, there is scope for integrating automated personality detection in almost all agents dealing with human-machine interaction such as voice assistants, robots, cars, etc. Research in this field is moving from detecting personality solely from textual data to visual and multimodal data. 

We have seen the practical potential for automated personality detection is huge in the present day scenario. We expect that many of the deep learning models discussed in this paper will be implemented for industrial applications in the next few years. For training these models there will be a need for large publicly available datasets of various personality measures. In the last 5 years, we have seen a tremendous rise in the number of deep learning models that have been used for automated personality detection. The best performance in most modalities (especially visual) have been achieved by deep neural networks. We expect this trend to continue in the future with smarter and more robust deep learning models to follow.  

\section{Conclusion} \label{Conclusion}
As discussed in this paper, there are many diverse applications of automated personality detection which can be used in the industry, hence making it a very hot and upcoming field. However, machine learning models are as powerful as the data used to train them. For a large number of cases in this field, enough labelled data are not available to train huge neural networks. There is a dire need of larger, more accurate and more diverse datasets for personality detection. Almost all of the current datasets focus on the Big-Five personality model and very few for other personality measures such as the MBTI or PEN. Normally, personality is measured by answering multiple questions in a survey. Assuming that everyone taking the survey answers honestly, the credibility of this survey in correctly labelling an individual's personality is still in question. A more accurate and efficient way of labelling personality traits needs to be explored. Most of the current methods for creating personality detection datasets rely on manual annotation through crowd sourcing using Amazon Mechanical Turk.

Recent multimodal deep learning techniques have performed well and are starting to make reliable personality predictions. Deep learning offers a way to harness the large amount of data and computation power at our disposal with little engineering by hand. Various deep models have become the new state-of-the-art methods not only for personality detection, but in other fields as well. We expect this trend to continue with deeper models and new architectures which are able to map very complex functions. We expect to see more personality detection architectures that rely on efficient multimodal fusion.

\section{Acknowledgement}
We would like to thank Prof.~Bharat M Deshpande for his valuable guidance. A.~Gelbukh recognizes the support of the Instituto Politecnico Nacional via the Secretaria de Investigacion y Posgrado projects SIP 20196437 and SIP 20196021.

\bibliographystyle{spmpsci} 
\bibliography{mynewbib}
\end{document}